\documentclass{article}
\usepackage{spconf,amsmath,graphicx}
\usepackage{enumitem}
\usepackage{epsfig}
\usepackage{graphicx}
\usepackage{amsmath}
\usepackage{amssymb}
\usepackage{amssymb}
\usepackage{pifont}
\usepackage{tabularx,multirow,booktabs}
\usepackage[ruled,vlined]{algorithm2e}
\usepackage{multirow}
\usepackage{enumitem}
\usepackage{subcaption}

\title{Self-improving object detection via disagreement reconciliation} 

\name{Gianluca Scarpellini$^{1,2}$, Stefano Rosa$^{1}$, Pietro Morerio$^{1}$, Lorenzo Natale$^{1}$, Alessio Del Bue$^{1}$ 
}
\address{$^{1}$Istituto Italiano di Tecnologia, Italy\\
$^{2}$University of Genova, Italy\\\tt\small \{name.surname\}@iit.it}
\begin{document}
\ninept
\maketitle
\begin{abstract}
Object detectors often experience a drop in performance when new environmental conditions are insufficiently represented in the training data. This paper studies how to automatically fine-tune a pre-existing object detector while exploring and acquiring images in a new environment without relying on human intervention, i.e., in a self-supervised fashion. In our setting, an agent initially explores the environment using a pre-trained off-the-shelf detector to locate objects and associate pseudo-labels. By assuming that pseudo-labels for the same object must be consistent across different views, we devise a novel mechanism for producing refined predictions from the consensus among observations. Our approach improves the off-the-shelf object detector by $2.66\%$ in terms of mAP and outperforms the current state of the art without relying on ground-truth annotations.
\end{abstract}
\begin{keywords}
Object detection, self-training, online exploration \end{keywords}
\section{Introduction}
The deployment of object detectors, particularly in real-world scenarios such as indoor robots, often requires fine-tuning with a custom dataset to accommodate for the domain shift between the training and test environments. While fine-tuning leads to improved performance, constructing a dataset can be a daunting task, requiring extensive time and effort in collecting and annotating thousands of images.

What if, instead of relying on human annotation, we could use data collected autonomously by the robot itself to fine-tune the detector? This paper explores this possibility by utilizing an embodied agent that actively explores an environment and leverages the collected data to improve object detection models. Object detectors play a crucial role in goal-oriented exploration \cite{chaplot2020object,cartillier2020semantic,chaplot2020learning}, and recent studies have demonstrated the relationship between object detection improvements and the reliable semantic mapping of environments \cite{chaplot2020object}, making them of significant importance in the field of robotics exploration.

This study draws parallels with the self-training paradigm \cite{lee2013pseudo}, where a model is trained on both labeled and unlabeled data. In self-training, a pre-trained model provides pseudo-labels for the unlabeled data, which may be refined by the introduction of additional noise \cite{arazo2019,xie2020selftraining,chen2020big}. Chaplot et al. \cite{chaplot2021seal} introduced a scenario where an agent explores the environment and collects data autonomously, proposing an approach to collect consistent pseudo-labels in the environment. While their approach effectively builds a semantic map and re-projects it onto each frame, it does not consider the inconsistency of detections as a viable training signal. Our approach, in contrast, presents a novel mechanism that leverages different points of view of the same object as a data augmentation strategy.

\begin{figure}[ht]
\begin{center}
\includegraphics[width=\linewidth]{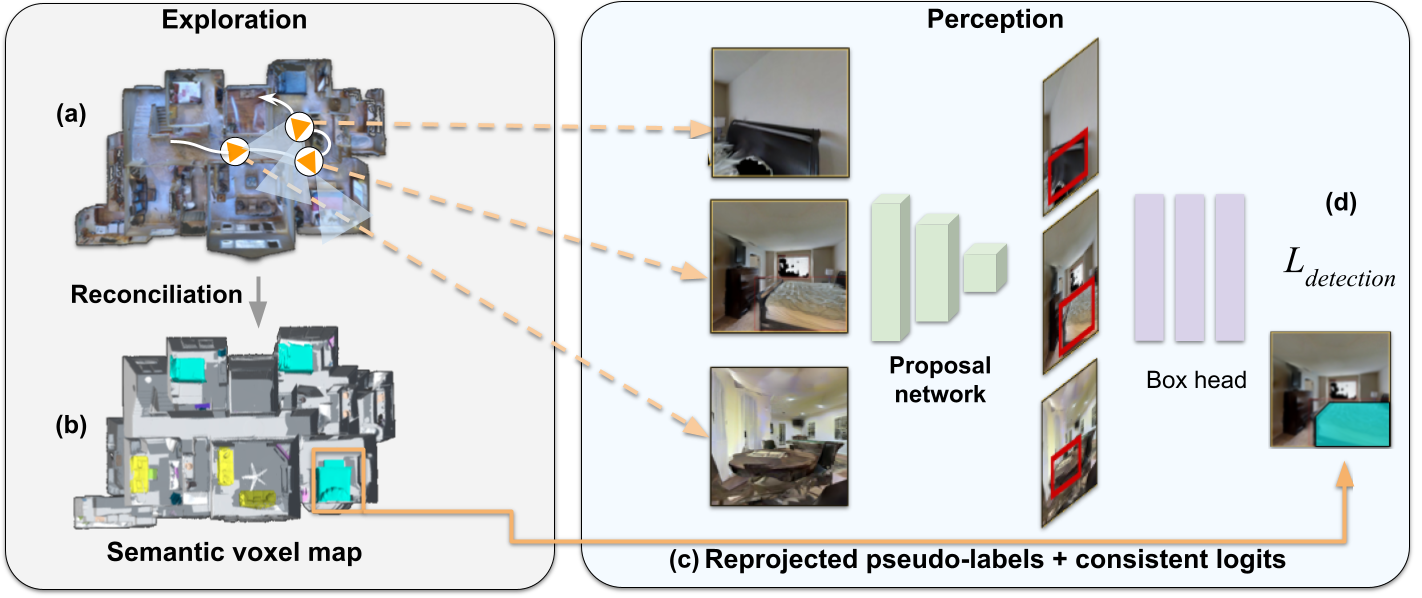}
   \caption{We equip an agent with an off-the-shelf MaskRCNN \cite{he2018mask} detector. \textbf{(a)} The agent explores a new environment and collects a set of noisy detections following a policy $\pi$ and builds an inconsistent voxel map. \textbf{(b)} we solve the inconsistencies in the voxel map and \textbf{(c)} project the reconciled semantic voxel map onto each observation to build the pseudo-labels for the self-training scheme. \textbf{(d)} Finally, we fine-tune the object detector relying only on the pseudo-labels.}
   \vspace{-20pt}
   \label{fig:fig1}
\end{center}
\end{figure}

Figure \ref{fig:fig1} summarizes our proposal. We implement our approach by equipping an agent with a pre-trained object detector, in this case MaskRCNN \cite{he2018mask}. The agent then explores an unknown environment and detects objects within the scene. The resulting predictions are used as pseudo-labels to fine-tune the detector. This process is conducted in three stages. Firstly, we enforce consistency of the predictions across multiple views through a consensus mechanism. The refined pseudo-labels are assigned to all relevant images and the consensus is used to fine-tune the detector. Secondly, using a contrastive learning approach, the model backbone is forced to map different views of the same object close together in the feature space while different instances are kept distinguishable. Lastly, we leverage the soft targets obtained by averaging multiple predictions for each instance as further feedback, to fully utilize the multiple acquired views.
In our approach, we utilize a contrastive loss function \cite{facenet2015} to ensure that features of an object are similar across its different views while features of different objects are distinguishable. Our method can be considered a generalization of the cropping strategy that is commonly utilized in self-supervised learning \cite{chen2020simple}. However, it should be noted that the cropping strategy alone is not sufficient to fully capture the intricacies of a 3D object viewed from various perspectives.

Our contributions are the following:
\begin{itemize}[topsep=0pt,parsep=0pt,partopsep=0pt]
 \setlength\itemsep{0em}
\item A technique to solve inconsistent pseudo-labels gathered during exploration;
\item A self-training module to improve the object detector adopting inconsistency and contrastive information;
\item A comparison between learned and hard-coded policies.
\end{itemize}

\section{Methodology}
\label{sec:methodology}

We equip a simulated agent with an object detector, a RGB-D sensor, and a position sensor. The agent \textit{looks around} and collects samples from a set of simulated environments \cite{xiazamirhe2018gibsonenv} in the Habitat simulator \cite{szot2021habitat}. A policy chooses actions for the agent among \textit{moving forward}, \textit{turn left}, and \textit{turn right}. We aim to fine-tune the agent's object detector by relying only on the samples it generates autonomously, i.e. without ground-truth annotations. 
As in previous literature \cite{chaplot2020semantic,chaplot2021seal}, we adopt the object detector MaskRCNN \cite{he2018mask} with a Resnet50 backbone \cite{resnet}; the model is provided by \cite{wu2019detectron2} and is pre-trained on COCO dataset \cite{lin2014microsoft}.
During the perception phase, we adopt the policy $\pi$ to collect a sequence of states $S_{0:N}$ for each environment. Each state $S_i$ contains the input RGB-D image $x_i$ and MaskRCNN's predictions $D_i$---a set of bounding-boxes $b^i_j \dots b^i_n$, instance segmentation masks $m^i_1 \dots m^i_n$, and normalized logits $\lambda_1 \dots \lambda_n$. Then, we build a semantic 3D representation of the environment. We propose a novel mechanism for producing consistent pseudo-labels across different views of the same environment while injecting the original inconsistencies as a soft-target to further help training. Finally, we leverage the rich information contained in our semantic point-cloud to fine-tune the MaskRCNN model. Next, we discuss the details for the pseudo-labels cleaning and the self-training method.

\subsection{Pseudo-labels and self-training} 
\label{sec:perception}
Pseudo-labels collected by an agent while navigating in a new environment might be very noisy, and instances may missed altogether if the object is seen from an unusual point of view. These inconsistencies have to be dealt with during self-training.

\begin{figure}[tb]
\begin{center}
\includegraphics[width=\columnwidth]{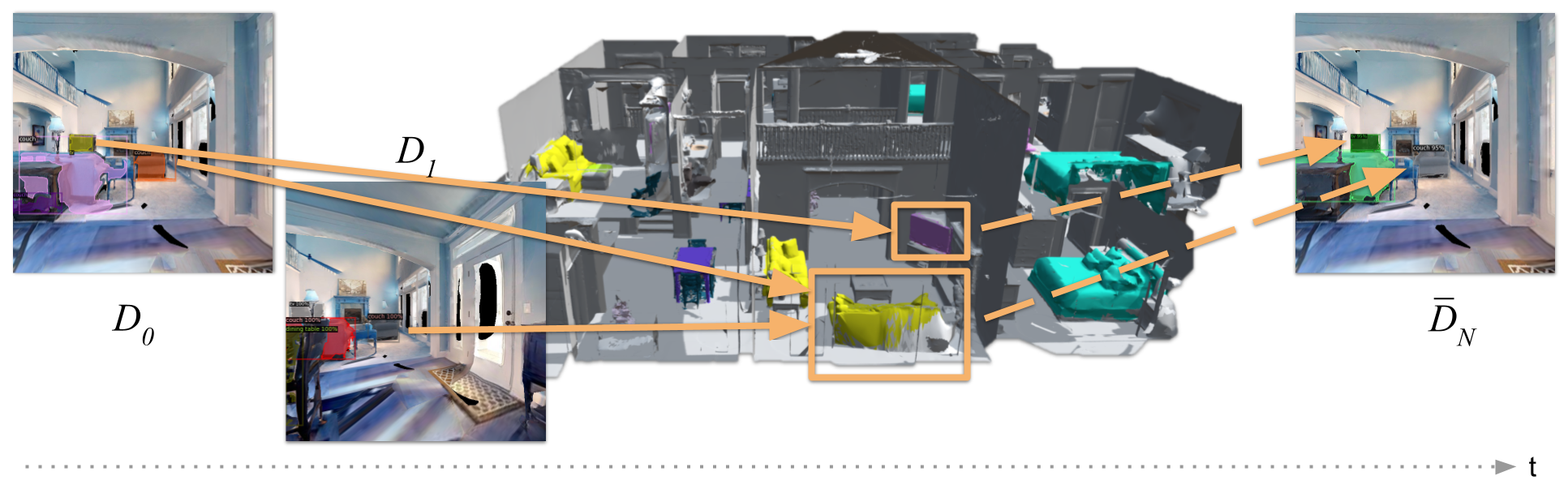}
 \vspace{-30pt}
\end{center}
  \caption{Semantic voxel map creation and projection of detections onto 2D frames. First, we aggregate detections $D_0, \dots, D_N$ into a semantic voxel-map. We solve the inconsistency of the voxel-map by assigning to each voxel the class with the maximum score among the predictions of the voxel. Next, we project the semantic voxel-map onto each observation, obtaining $\overline D_0, \dots, \overline D_N$. $\overline D_N$ is the consistent pseudo-label for observation $N$ and is obtained by reprojecting the voxel-map onto RGB-D frame $x_N$. Each pseudo-label $\overline D_i$ is associated to an object instance via the identifier $u_i$ and contains the consistent logits vector $\overline \lambda_{u_i}$.}
  \label{fig:reproject}
  \vspace{-10pt}
\end{figure}

We propose to leverage tri-dimensional projection to build a consistent voxel map from noisy predictions as summarised in Figure \ref{fig:reproject}. Building upon the standard voxelization process, we introduce a consensus mechanism to improve the quality of the voxel map. We project the voxel map back onto each observation to obtain consistent pseudo-labels for our self-training scheme. 

During the exploration phase, we exploit the depth and position sensors of the agent to build a semantic point cloud by projecting the agent's detector predictions $D_i=(b_i, m_i, y_i, \lambda_i)$ (respectively, instance segmentation masks, classes, and logits from the 2D detector) to 3D coordinates. First, we voxelize the resulting point-cloud with $\mathtt{voxel\_size}=0.05m$. In the resulting voxel map, each non-empty voxel maintains the set of all logits vectors. This results in an inconsistent map, where a voxel is assigned to possibly multiple classes. For this reason, we process the voxel map by keeping, for each voxel, a consistent hard-label $\overline y$. 

We then compute the hard label $\overline y$ of a voxel as the class with the maximum score among all the predictions associated to that particular voxel. We highlight however that this voxelization process is \textit{lossy}, as we compute $\overline y$ via a \textit{max} operator.
We aggregate connected voxels with equal classes $\overline y$ via a 26 connect-components algorithm \cite{cc3d}. Each aggregated component is a distinct object instance, therefore it is uniquely assigned an identifier $u$, and has a unique class (the same as the aggregated voxels). 

The resulting voxel map is \textit{consistent}, as every object instance has a unique class $\overline y$. As previously discussed, this process drops important information about the inconsistencies between the predictions. To overcome this issue, we compute for each object instance $u_i$ a \textit{consistent} object logits $\overline \lambda_{u_i}$ as:
\begin{equation}
\label{eq:clean}
\overline \lambda_{u_i} = \frac{1}{| \mathcal{Q}(u_i)|}\sum_{\lambda_j \in Q(u_i)} \text{Softmax}( \lambda_j ),
\end{equation}
where $\mathcal{Q}(u_i)$ is the set of predicted vector logits $\{\lambda_i\}_i$ for the object instance $u_i$ associated to detection $i$. We further discuss the consistent vector logits in the following Sections, where we underline the importance of this soft-target for the fine-tuning of the detector.

Any 3D object instance $u$ that appears in the semantic map is projected onto each $x_i$ using the camera intrinsics and extrinsics matrix, resulting in a per-instance semantic mask $\overline m_i$ juxtaposed onto $x_i$. We compute the bounding-boxes $\overline b_i$ as the minimum rectangle containing $\overline m_i$ and assign the object's class, consistent logits, and identifier to the pseudo-label. 
The projection phase solves both issues of the agent's predictions: \textit{(i)} as the voxel map is self-consistent, it follows that pseudo-labels for an object instance are consistent across different views \textit{(ii)} the consistency also applies for frames where no detections were obtained at first.  
Our training set $\mathcal{D}$ consists of the resulting set of observations and pseudo-annotations $\{x_i, (u_i, \overline \lambda_{u_i}, \overline m_i, \overline b_i, \overline y_i)\}_i$. 
\\
\textbf{Self-training phase.}
We propose a fine-tuning strategy for MaskRCNN's head that leverages the consistent pseudo-labels described above. Given the training set $\mathcal{D}$, we first compute mask, bounding-boxes, and classes losses $\mathcal L_\text{head}$ on pseudo-labels $\overline D_0, \dots \overline D_N$ as in \cite{he2018mask}. To fully exploit the information contained in our semantic voxel map, we propose two additional steps in the self-training strategy: instance-matching and soft distillation. 
\begin{figure}[tbh]
\begin{center}
\includegraphics[width=\linewidth]{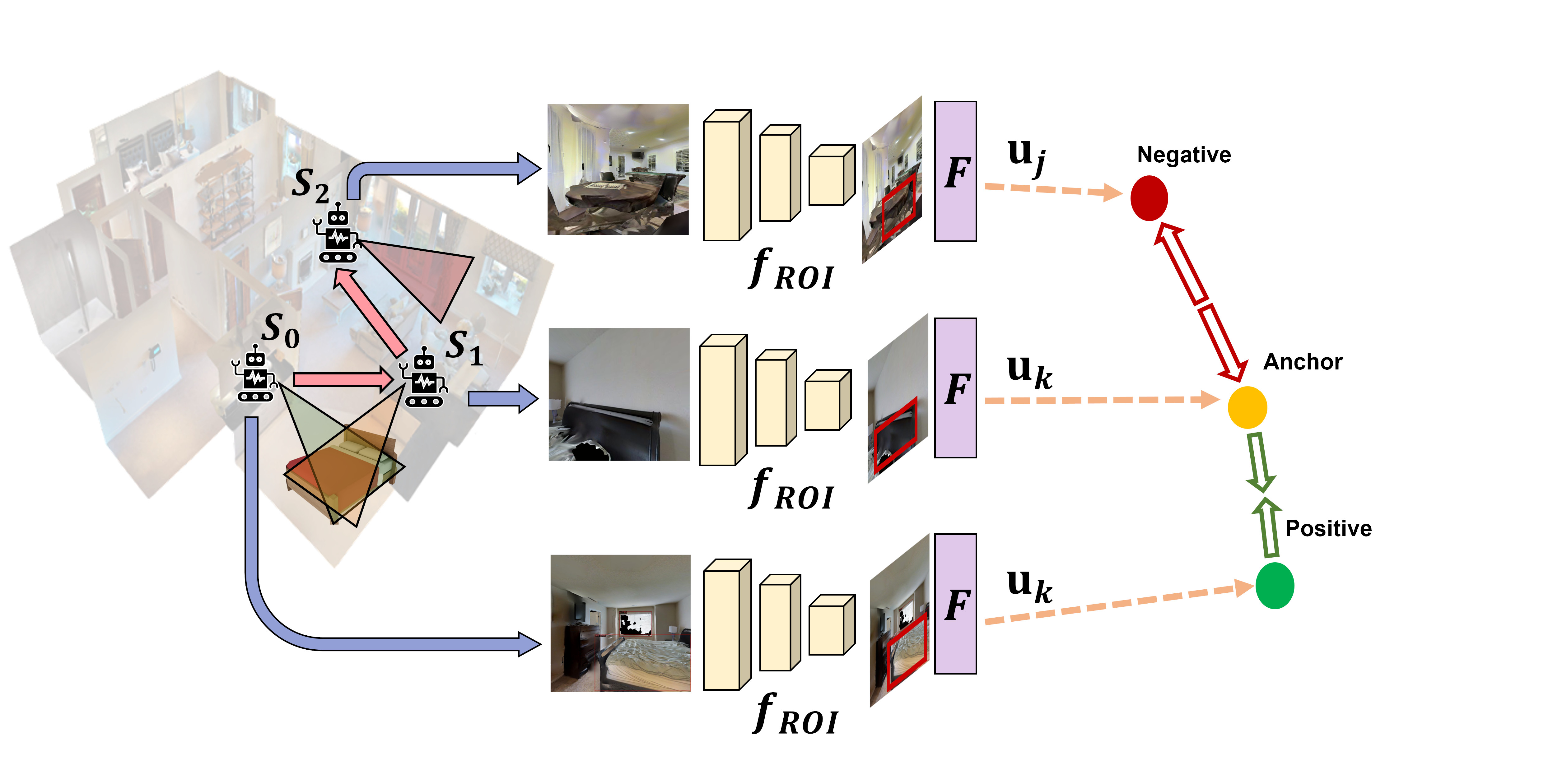}
 \vspace{-30pt}
\end{center}
  \caption{The instance-matching loss exploits disagreements between predictions for the same object. In fact, it enforces feature vectors belonging to the same object (e.g., $\text{u}_k$ in the Figure) to be close in the feature space, while moving away feature vectors of different objects (e.g., $\text{u}_j$ in the Figure).}
  \label{fig:triplet}
   \vspace{-10pt}
\end{figure}
\\
\textbf{Instance-matching.}
The first component of our strategy is a contrastive triplet loss between detections. Intuitively, a contrastive loss helps $f_\text{box-head}$ construct an embedded feature-space where feature vectors of the same objects are closer than feature vectors of different objects, as shown in Figure \ref{fig:triplet}. If an object is seen from extremely diverse and challenging points of view, nearing the respective feature vectors is beneficial for the model's feature projector.
Given a batch of observations $x_i$, we predict the regions of interest $\text{ROIs}_{i}$ through $f_\text{ROI}$ of MaskRCNN. Next, we provide the ROIs to the model's head $f_\text{box-head}$ to predict bounding-boxes $B \in \mathbb{R}^{k\times4}$, per-class logits $\lambda \in \mathbb{R}^{k\times |C|}$, and a set of feature vectors $F \in \mathbb{R}^{k\times1024}$. 

For each observation, we associate the predictions with the corresponding pseudo-labels and assign their object instance identifier to each feature vector. Next, for a given set of feature vectors $F$, we compute positive and negative relationships by leveraging the object identifiers and finally compute $\mathcal L_\text{im}$ by taking the distance between positive and negative examples.
\\
\textbf{Consistent logits as soft pseudo-labels.}
As discussed in Section \ref{sec:perception}, we expect that the model can benefit from an additional loss that guides the prediction towards a smooth average of the logits provided by the off-the-shelf detector from different views. The knowledge distillation loss--our second component--introduces consistent logits as a form of soft distillation. We compute $\mathcal L_\text{distill}$ as the cross-entropy between the predicted logits $\lambda_i$ and the consistent logits $\overline \lambda_{u_i}$.
\\
\textbf{Training.} To summarize, we propose the combined loss $\mathcal L_\text{detection}$ as:
\begin{equation}
\label{eq:detection}
\mathcal L_\text{detection}= \mathcal L_\text{im} + \alpha \mathcal L_\text{distill} + \mathcal L_\text{head},
\end{equation}
by merging three different losses: \textit{(i)} a triplet loss based on different views of the same object $\mathcal L_\text{im}$, \textit{(ii)} a soft distillation loss $\mathcal L_\text{distill}$ for leveraging \textit{consistent} logits $\overline \lambda$, and \textit{(iii)} mask, bounding-boxes, and classes losses $\mathcal L_\text{head}$ on pseudo-labels.
We proceed to fine-tune the model by jointly minimizing $\mathcal{L}_\text{detection}$, and the standard region proposal loss $ \mathcal{L}_\text{rpn}$  introduced in \cite{ren2016faster}. Algorithm \ref{alg:finetuning} summarizes the proposed method.

\begin{algorithm}[tb]
\small
 Initialize observation queue $Q$, detector's box-head parameters $w$, policy $\pi$;
 \For{each training environment}{
  Collect a trajectory $S_0,  \dots S_{N-1} $ following policy $\pi$ \;
  Build semantic point-cloud from predictions $D_{0:N-1}$, depth information, and position sensor readings\;
  Obtain a consistent point-cloud via voxelization and 3d connected components\;
  Produce hard labels, soft labels, and unique object identifiers $\overline D_{0:N-1}$ by raycasting the consistent point-cloud onto each observation $x_{0:N-1}$\;
}
  \For{K epochs}{
    Sample a mini-batch from Q \;
    Predict the region proposals and compute $\mathcal L_\text{rpn}$\;
    Predict classes, bounding-boxes, and semantic masks, and compute $\mathcal L_\text{head}$\;
    For each class prediction, compute $\mathcal L_\text{distill}$ by comparing the predicted logits $\lambda_i$ with the consistent logits $\overline \lambda_{u_i}$\;
    For each bounding-box prediction, extract the associated feature-vector and compute $\mathcal L_\text{im}$, as in Figure \ref{fig:triplet}\;
    Compute $\mathcal L_\text{detection}$ in Equation \ref{eq:detection}\;
    Aggregate $\mathcal L_\text{rpn}$ and $\mathcal L_\text{detection}$ and update object detector parameters $w$\;
 }
 \caption{Our proposed self-training scheme. The perception phase relies only on agent sensors and on a off-the-shelf object detector. No ground-truth labels are required.}
 \label{alg:finetuning}
\end{algorithm}
\section{Experiments}
\label{sec:experiments}
We devise a set of experiments to test policies and fine-tuning techniques separately. 
For each policy described in Section \ref{sec:policies}, we let an agent move in the environment and collect 7,500 frames from the Gibson dataset \cite{xiazamirhe2018gibsonenv}. We opt for MaskRCNN \cite{he2018mask} as the object detector as in previous approaches, although our proposed approach is model-agnostic.
\\
\textbf{Training details} We adopted the MaskRCNN model \cite{he2018mask} provided by Detectron2 \cite{wu2019detectron2} and pre-trained on the COCO dataset \cite{lin2014microsoft}. MaskRCNN provides bounding-boxes and masks for each prediction. We set the score threshold for object-detection predictions at $0.7$. Similarly to previous works \cite{chaplot2020semantic,chaplot2020object}, we focus on a subset of objects which are common in houses and living spaces: \textit{toilet}, \textit{couch}, \textit{bed}, \textit{dining table}, \textit{potting plant}, and \textit{tv}. 
For the self-training stage, we adopt SGD optimizer with learning rate $\text{lr}=1e-4$, $\text{epochs} = 10$, $\text{weight-decay} = 1e-5$, $\text{momentum} =0.9$ and $\text{batch size} = 16$. Intuitively, the soft-distillation loss $\mathcal L_\text{distill}$ should not dominate over cross-entropy, thus leading to incompatibilities between the two. We choose to set the soft-distillation weight $\alpha=0.7$ by empirically comparing the gradients of the two losses. We conduct additional ablation experiments in Section \ref{sec:ablations} to study the impact of soft-distillation on the training. 
For the triplet-loss, we adopt the default $\text{margin}=0.3$.
\\
\textbf{Evaluation} To evaluate our approach, we sample 1,000 observations from each test scene, which results in 4,000 testing samples. The test set annotations are provided by Habitat and originally proposed in Armani et al. \cite{armeni20193d}. We adopt the mean Average Precision with IOU threshold on bounding-boxes at 0.5 as evaluation score \cite{lin2014microsoft}. We stress that no semantic annotations are required during training. We introduce the exploration policies in Section \ref{sec:policies} and discuss the baseline perception methods in Section \ref{sec:perception}. Section \ref{sec:results} compares our self-training approach with the baselines. 
\subsection{Exploration methods}
\vspace{-5pt}
\label{sec:policies}
We adopt the following classic and learned policies for exploring the environment: \\  
\textit{\textbf{Random goals}} -- the agent chooses a feasible random goal in the map. Next, the agent moves to the selected point via a global path planner. \\ 
\textit{\textbf{Frontier exploration}} \cite{yamauchi1998frontier} -- implements a simplified version of classical frontier-based active exploration \cite{yamauchi1998frontier}. The agent keeps an internal representation of the explored map, computes goal points of interest at the frontiers of the explored map, selects the next goal in a greedy manner, and moves towards it via a global path planner. \\
\textit{\textbf{NeuralSLAM}} \cite{chaplot2020learning} -- RL policy based on long-term and short-term goal predictions for maximizing map coverage. A global policy predicts the next long-term goal, while a local policy predicts a sequence of short-term steps. The reward is the percentage of exploration of the environment. \\
\textit{\textbf{Semantic Curiosity}} \cite{chaplot2020semantic} -- RL exploration policy that predicts local steps to maximize inconsistency between detections projected onto the ground plane. \\
\textit{\textbf{SEAL}} \cite{chaplot2021seal} -- RL policy for maximizing the number of confident predictions of the detector. The agent builds a voxel map of the environment by leveraging the sensor information. A global policy processes the current voxel map and position information and predicts the next long-term goal, while a local path-planner guides the agent toward the goal. The reward is the number of voxels assigned to any possible class with a score above $0.9$. We implement \cite{chaplot2020semantic} and \cite{chaplot2021seal} to the best of our efforts since no public implementation is available.

\subsection{Perception methods}
We compare our \textit{perception module} with two baselines:\\
\textit{\textbf{Self-training}} \cite{yalniz2019billion} -- the perception phase consists in adopting the prediction of the off-the-shelf object detector as pseudo-labels for training, without any further processing. We note that self-training often requires several millions of images to provide strong benefits. \\
\textit{\textbf{SEAL perception}} \cite{chaplot2021seal} -- SEAL aggregates the predictions of the off-the-shelf object detector into a semantic voxel map by assigning to each voxel the class with maximum score among the predicted classes. Then, the voxel-map back is projected onto each observation. Finally, the detector is fine-tuned by minimizing the classical MaskRCNN losses \cite{he2018mask}.\\
\textbf{\textit{Ours}} -- our perception module applies multi-view consistencies among predictions while retaining the inconsistency information as soft-target and imposing feature vectors of the same object to be close when the object is seen from different views, as discussed in Section \ref{sec:perception}.

\subsection{Results}    
\label{sec:results}
We evaluate our method against two perception methods for all the exploration policies. Table \ref{tab:results_policies} reports our results. As a reference, we assess that the off-the-shelf MaskRCNN \cite{he2018mask} reaches mAP@50 $40.33$ on our test set. 
Our perception method outperforms self-training \cite{yalniz2019billion} and SEAL's perception \cite{chaplot2021seal} in combination with random, frontier, Semantic Curiosity, and NeuralSLAM baselines. Even when using the random goals baseline, therefore without relying on any predetermined behavior, our perception module improves over self-training by $2.21\%$. When compared with exploration policies, both learned \cite{chaplot2020learning} and classical \cite{yamauchi1998frontier} ones, our perception module outperforms SEAL perception. In particular, we notice that a classical frontier exploration baseline outperforms NeuralSLAM and SEAL in our self-learning task.  
Intuitively, if the variance of the observations is low, the semantic voxel map reflects the labels predicted by the off-the-shelf detector. 
NeuralSLAM exploration collects fewer observations for each object instance. In fact, all the perception methods reach similar mAP when NeuralSLAM is adopted. Both SEAL and our perception modules are able to solve the inconsistencies. SEAL perception slightly outperforms our perception when applied in conjunction with SEAL policy. As SEAL agents find views of an object with the highest prediction score, thus limiting the number of views per object and, consequently, the performances of our method. Indeed, we argue this is not surprising, since these two components were carefully crafted to work jointly together. Semantic Curiosity predicts only local actions, which limits the capacity of the policy to explore the environment and, therefore, the collection of training data.
On the other hand, frontier exploration combined with our perception module achieves an overall highest AP of $43.09$. Our intuition is that, due to the greedy exploration policy, the agent tends to move through the same areas multiple times, collecting more views of the same object, which is then cleaned by the consensus approach. 

\begin{table}[tb]
\caption{\label{tab:results_policies}We compare the mAP of our perception phases with the baselines. Our approach outperforms the perception baselines in almost all scenarios and reaches the overall mAP@50 $43.09$ with the classical frontier exploration. As a reference, off-the-shelf MaskRCNN reaches mAP@50 40.33.}
\centering
\resizebox{\columnwidth}{!}{
\begin{tabularx}{\columnwidth}{l|ccc}
\hline
\hline
\multirow{2}{*}{Policy} & Self-training & SEAL & \textbf{Our} \\
& \cite{yalniz2019billion} & perc. \cite{chaplot2021seal} & \\
\hline\hline
Random goals & 39.67 &41.19& \textbf{41.88}\\ 
Frontier \cite{yamauchi1998frontier} &  40.18 & 41.98 & \textbf{43.09 }\\  
NeuralSLAM \cite{chaplot2020learning} & 39.98 & 39.56&\textbf{40.32}  \\  
Sem. Curiosity \cite{chaplot2020semantic}&40.23&41.06&\textbf{41.37}\\
SEAL policy \cite{chaplot2021seal} & 39.33 & \textbf{43.01} & 42.38\\ \hline

\end{tabularx}
}

\end{table}

\begin{table}[tb]
\caption{\label{tab:abl}Impact of $\mathcal{L}_{im}$ and $\mathcal{L}_{distill}$. Both losses improve results significantly if data coverage is enough, as in frontier exploration.}
\centering
\begin{tabular}{p{2.1cm}|p{0.5cm}p{0.6cm}|p{0.6cm}p{0.6cm}p{0.6cm}p{0.6cm}}
\hline
\hline
    \multirow{2}{*}{Policy} &\multicolumn{2}{c}{$\mathcal{L}_{im}$ } &\multicolumn{4}{c}{$\alpha \mathcal{L}_{distill}$ }\\
 & \ding{55} & \ding{51} & \ding{55}  & \textbf{0.1} & \textbf{0.7} & \textbf{1.0}\\
 \hline
Random & 41.58& 41.88   & 41.01&41.74& 41.88 &41.32\\
Frontier \cite{yamauchi1998frontier}&42.32&  43.09 &42.23&42.29&  43.09&43.08\\
NeuralSL. \cite{chaplot2020learning} &40.05&  40.32  &38.70&39.65&  40.32&40.26\\
Sem. Cur. \cite{chaplot2020semantic}&42.01&41.37&43.16&42.60&41.37&41.25\\
SEAL \cite{chaplot2021seal}&42.35&  42.38&42.72& 42.28& 42.38 &41.16\\
\end{tabular}
\vspace{-10pt}
\end{table}

\textbf{Ablations}
\label{sec:ablations}
We study the impact of our instance-matching loss and consistent logits for reaching the best performance.\\
\textit{Instance matching.} 
We expect our instance-matching method to benefit from an exploration policy that increases the number of views for an object. Table \ref{tab:abl} reports the results of applying instance-matching in conjunction with different policies. We notice that the object detector benefits from the contrastive approach in conjunction with all the policies except for semantic curiosity. In particular, we highlight an overall improvement of $.77\%$ for frontier exploration. 
\\
\textit{Distillation loss.} We ablate the contribution of the aggregated soft consistent logits by comparing different values of $\alpha$, which regulates the weight of the distillation loss $\mathcal L_\text{im}$ and thus the training signal derived by injecting the consistent logits $\overline \lambda$ with a soft-distillation loss. All policies, except for SEAL and Semantic Curiosity, improve their performance if we introduce $\mathcal{L}_\text{distill}$, even when its weight is only $\alpha=0.1$. As discussed in Section \ref{sec:perception}, we expect the soft-distillation to improve the training when it does not exceed the cross-entropy loss for hard pseudo-labels. We find that weighting $\mathcal{L}_\text{distill}$ with $\alpha=0.7$ is the best option: \textbf{frontier + our perception } outperforms the other baselines and achieves the state-of-the-art result. On the other hand, $\mathcal{L}_\text{distill}$ damages the results when backed by SEAL or Semantic Curiosity policies. This result is expected. As discussed in Section \ref{sec:experiments}, policies with limited variance or training data cannot benefit from injecting the consistent logits through $\mathcal{L}_\text{distill}$.

 \section{Conclusion}
We proposed a novel self-training mechanism for object detection that relies on the exploration of the environment. The object detector can benefit from hard positive examples, e.g. where the off-the-shelf detector fails. Namely, the perception phase can exploit disagreements among pseudo-labels. 

We propose to clean the inconsistencies by using a semantic voxel map by accumulating each prediction into the 3D space. We solve disagreeing labels in a 3D voxel space via a novel mechanism, while preserving the conflicting information among predictions. By projecting the 3D volumes onto each training examples we generate a new training set. Next, we fine-tune the object detector by leveraging the resolved the pseduo-labels as well as consistent logits vector for each object. Finally, we force the feature vector of each object to be consistent across different views. Experiments show the benefits of our perception stages which outperform the baselines and reaches state-of-the-art performance, improving the off-the-shelf object detector by $2.76\%$.

\bibliographystyle{IEEEbib}
\bibliography{egbib}

\end{document}